\documentclass[5p,sort,compress]{elsarticle}
\usepackage{fancyhdr}
\usepackage{amsfonts}
\usepackage{url}
\usepackage{amsmath}
\usepackage[labelfont=bf,justification=raggedright,singlelinecheck=false]{caption}
\usepackage{array}
\usepackage{multirow}
\usepackage{graphicx}
\usepackage{amsmath}
\usepackage{algorithm}
\usepackage[noend]{algpseudocode}
\usepackage{pdflscape}
\usepackage{rotating}
\usepackage{wrapfig}
\usepackage[inline, shortlabels]{enumitem}
\usepackage{graphbox}
\usepackage{mathtools}
\usepackage{amsmath}

\newcolumntype{v}{>{\centering\arraybackslash}m{.18\linewidth} }
\usepackage{hyperref}
\hypersetup{
	colorlinks=true,
	linkcolor=blue,
	filecolor=magenta, 
	urlcolor=cyan,
}
\usepackage[labelfont=bf, justification=justified, format=plain]{caption} 

\begin{document}
	\journal{Arxiv}
	\title{An Effective Scheme for Maize Disease Recognition based on Deep Networks}
	\author[fum1]{Saeedeh Osouli}
	\ead{saeedeh.osouli@mail.um.ac.ir}
	\author[fum2,MVLAB]{Behrouz Bolourian Haghighi\corref{cor1}}
	\ead{b.bolourian@mail.um.ac.ir}
	\author[fum3]{Ehsan Sadrossadat}
	\ead{ehsan.Sadrossadat@research.uwa.edu.au}
	\cortext[cor1]{Corresponding author}
	\address[fum1]{Electrical Engineering Department, Ferdowsi University of Mashhad, Mashhad, Iran }
	\address[fum2]{Computer Engineering Department, Ferdowsi University of Mashhad, Mashhad, Iran }
	\address[MVLAB]{Machine Vision Laboratory, Ferdowsi University of Mashhad, Mashhad, Iran}
	\address[fum3]{School of Engineering, University of Western Australia, Perth, Australia}
	\begin{abstract}
		In the last decades, the area under cultivation of maize products has increased because of its essential role in the food cycle for humans, livestock, and poultry. Moreover, the diseases of plants impact food safety and can significantly reduce both the quality and quantity of agricultural products. There are many challenges to accurate and timely diagnosis of the disease. This research presents a novel scheme based on a deep neural network to overcome the mentioned challenges. Due to the limited number of data, the transfer learning technique is employed with the help of two well-known architectures. In this way, a new effective model is adopted by a combination of pre-trained MobileNetV2 and Inception Networks due to their effective performance on object detection problems. The convolution layers of MoblieNetV2 and Inception modules are parallelly arranged as earlier layers to extract crucial features. In addition, the imbalance problem of classes has been solved by an augmentation strategy. The proposed scheme has a superior performance compared to other state-of-the-art models published in recent years. The accuracy of the model reaches 97\%, approximately. In summary, experimental results prove the method's validity and significant performance in diagnosing disease in plant leaves.
	\end{abstract}
	
	\begin{keyword}
		Maize Disease Detection\sep Deep Learning\sep Convolutional Neural Networks(CNN)\sep Transfer Learning\sep Image Classification.
	\end{keyword}
	\maketitle
	\begin{figure*}[t]
		\centering
		\includegraphics[width=1\textwidth,trim= 1cm 24cm 1cm 3cm,clip]{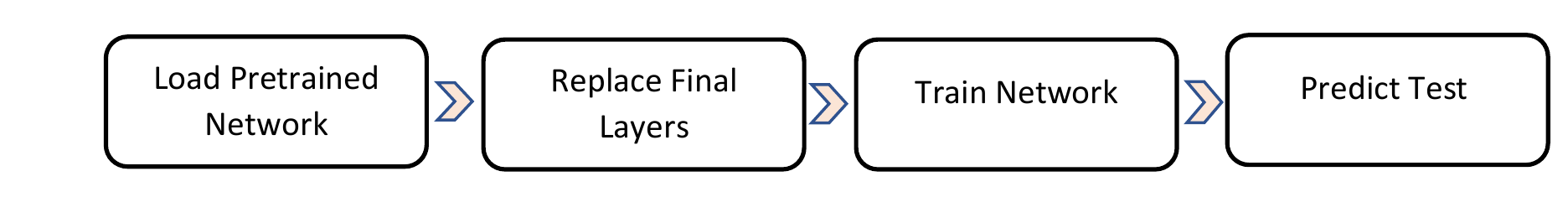}
		\caption{The block diagram of transfer learning strategy.}
		\label{fig:tl}
	\end{figure*}
	\section{Introduction}
	Maize is one of the oldest crops that is directly used as the primary food of numerous people, livestock, poultry, and in various industries, even energy production. The area under cultivation and the amount of maize production worldwide are constantly increasing. Due to its desirable characteristics, such as high adaptability, maize has a special place among crops. Detailly, its area under cultivation at the beginning of the current century was more than 120 million hectares of agricultural land globally. Generally, it dedicates third place after wheat and rice among various products.
	
	Overall, there are eight common diseases in these plants, including curvularia leaf spot, dwarf mosaic, gray leaf spot, northern leaf blight, brown spot, round spot, rust, and southern leaf blight \cite{ref1}. The mentioned diseases in plants and fruits significantly affect the quality and quantity of agricultural products. Accordingly, farmers suffer from growing problems with various types of plant diseases. It is worth mentioning that sometimes herbalists cannot diagnose diseases correctly. Hence, automatic detection of plant diseases is one of the essential factors in monitoring agricultural products and detecting timely diagnosis of disease in the leaves of plants.
	
	Regrettably, the mentioned diseases in the leaves of plants reduce the presentation of agricultural products. Therefore, punctual detection of plant diseases causes the necessary actions to prevent losses, improve plant quality, reduce the use of pesticides, and increase farmers' incomes. Nowadays, researchers have overcome the problems with the help of artificial intelligence science, such as neural networks and image processing\cite{ref2}. Although the early and timely prediction of disease diagnosis in plants has a remarkable impact on agricultural production management, the farmers in rural areas cannot easily access specialists for this aim. Actually, the visual observations are the primary approach of diagnosing plant leaf disease \cite{ref3}. This approach requires constant review by experts, which is extremely exhausting and costly \cite{ref4}. In the case of large-scale cultivation, this system can also be combined with automatic agricultural vehicles to detect diseases by continuously imaging plants \cite{ref5}. In summary, looking for a fast, automatic, less expensive, and accurate method to perform plant disease detection plays a crucial role in agricultural science\cite{ref6}.
	
	Report of Food and Agricultural Organization (FAO) estimated that the world population would reach 9.1 billion by 2050; thus, about 70\% growth in food production will be required for a steady supply \cite{ref7}. Nevertheless, performing traditional techniques to diagnose the disease in plants is not efficient and optimal. Because it sometimes causes tardiness or incorrect diagnosis of the disease. On the other hand, it leads to excessive use of pesticides which is more harmful and reduces the production of products \cite{ref8}. Recently, considerable advances have been made in the development of automated empowerment models and accurate and timely identification of plant leaf diseases \cite{ref5}. This is due to the promotion of technology in artificial intelligence, especially image processing, and providing reliable data and appropriate hardware.
	
	\begin{figure*}[t]
		\centering
		\includegraphics[width=1\textwidth,trim= 1.5cm 8cm 2cm 0cm,clip]{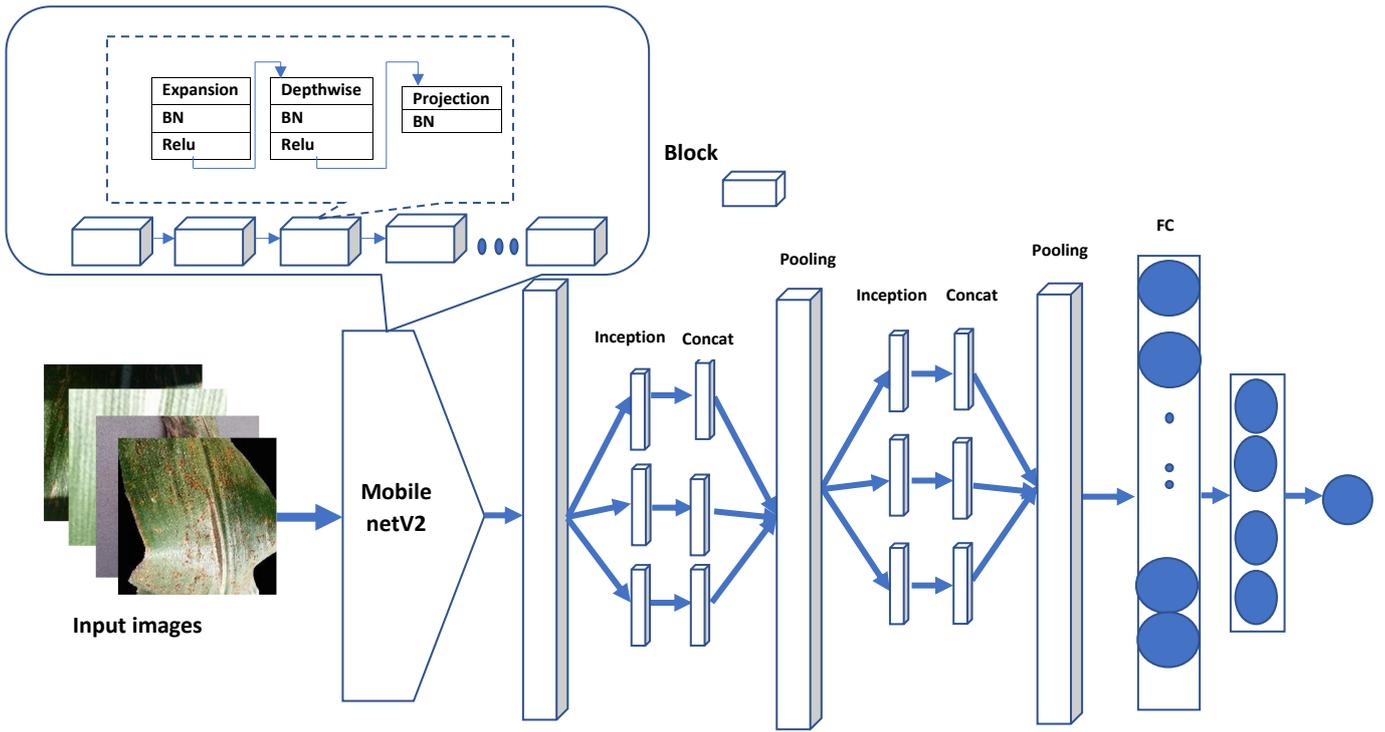}
		\caption{The structure of proposed network.}
		\label{fig:diagram}
	\end{figure*}
	\subsection{Literature Review}
	Most research classified plant images into only two categories in the last decade, including healthy and diseased. In other words, many previous works have taken into account the image binary classification to predict healthy or unhealthy instances. Moreover, some methods have been presented in agriculture to identify and diagnose plant diseases. For this aim, traditional machine learning approaches have been widely adopted.
	
	Generally, plant leaves are one of the essential sources for diagnosing the disease \cite{ref9}. The image processing and learning techniques take advantage of the significant features of leaves to diagnose plant diseases. These methods included Support Vector Machine (SVM), k-Nearest Neighbors( KNN), Fisher Linear Discriminant(FLD), Random Forest(RF), Artificial Neural Network (ANN), Gabor Transform, Global-Local Singular Value, and sparse representation, Scale-Invariant Feature Transform (SIFT), etc\cite{ref10, ref11, ref12, ref13, ref14, ref15, ref16}. In \cite{ref22} as research review work, Convolutional Neural Networks are introduced as one of the most popular techniques in the field of classification, especially for the image and video in agriculture. 
	
	After the presentation of the LeNet article in\cite{ref17}, the deep learning strategies, especially in the field of image classification, have made significant progress. In \cite{ref18}, authors used different pooling operations, filter sizes, and algorithms to identify ten common rice diseases. The proposed Convolution Neural Networks (CNNs) achieved an accuracy of around 95.48\%. Authors in \cite{ref19} used two deep learning architectures, including Alex and Google Networks, to diagnose and classify plant diseases on the plant village dataset. The training and test cases were divided into several categories in this work. The accuracy rate reached about 99/35\% when the data were split into 20\% for testing and 80\% for training. 
	
	In \cite{ref20} CNN layers are trained to identify the northern leaf blight of maize automatically. This approach demonstrated the challenge of the limited data and the numerous irregularities seen in farm plant images. The scheme generated an accuracy of around 96.7\%. In \cite{ref21}, the transfer learning approach was employed to classify samples. This is a process of using a pre-trained model (AlexNet) to classify new categories of the image, which is performed for disease classification in this work. The method was able to classify 26 types of diseases in 14 different types of plants with 99.35\% accuracy. In \cite{ref23}, the improved GoogLeNet and Cifar10 models based on deep learning are proposed for the identification of maize leaf diseases. These two enhanced models were used to test 9 kinds of maize leaf images and obtained admissible results.
	
	In \cite{ref5}, authors developed and studied two well-known deep learning architectures to diagnose diseases in plants. The training samples came from an open database of 87,848 leave images of healthy and diseased plants. The researchers in \cite{ref24} used deep learning architectures such as AlexNet and VGG16 networks to classify tomato crop diseases. In another work, \cite{ref25}, a comparison of the deep learning architectures was presented. Particularly, VGG 16, Inception V4, ResNet with 50, 101, and 152 layers, and DenseNets with 121 layers were evaluated as popular models in this work. The data which is used included 38 different classes of healthy and diseased plants for 14 plant species. In this study, the best accuracy for DenseNet architecture has been 99.75\%.
	
	Researchers in \cite{ref26} used the low-shot learning method to control and prevent disease in tea leaves promptly. In this study, the textual and non-textual characteristics were extracted from the images of tea plant leaves. In the following, the SVM algorithm was adopted for segmentation. In this way, the image's background is removed, and only the patient stain remains. After this step, the Conditional Deep Convolutional Generative Adversarial Networks (C-DCGAN) were used to increase the number of data. Finally, the data was provided to the VGG16 convolutional neural network to diagnose the disease in the tea plant leaves. The accuracy of the proposed method has also reached 90\%.
	
	In \cite{ref27}, instead of considering the whole leaf, the authors used individual lesions and spots to diagnose the disease. Because each area has its characteristics, the data change increases without the need for additional images. This has also made it possible to identify multiple diseases affecting the same leaf. It should be noted that the proper selection of signs must still be made manually, and complete automation must be avoided. The accuracy obtained in the proposed method is, on average 12\% higher than when taught only with the main images of the network. Also, it did not contain any product exactly below 75, even if ten diseases were considered.
	
	In \cite{ref28}, authors proposed the recognition and classification of maize plant leaf diseases by employing the Deep Forest algorithm. The proposed approach has outperformed the other traditional machine learning algorithms in terms of accuracy.
	
	In \cite{ref29}, the proposed model gives a comparative study of the above three deep learning models, including LeNet, AlexNet, and InceptionV3, for categorizing 38 varieties of plant diseases. It can be observed based on the results that InceptionV3 performs better with an accuracy of 98\% compared to other models. In \cite{ref30}, to diagnose the disease in the leaves of plants, researchers used a combination of two methods of convolutional neural networks and auto-encoders. In this work, a convolution filter with sizes 2$\times$2 and 3$\times$3 was used. In this way, 97.50\% and 100\% accuracy were achieved for filters with sizes 2$\times$2 and 3$\times$3, respectively. In \cite{ref6}, authors used transfer learning of the deep convolutional neural networks for the identification of plant leaf diseases and considered using the pre-trained model. The method reached effective performance with the help of this technique.
	
	\subsection{Key Contributions}
	The main purpose of this work is to design an accurate model for diagnosing disease in maize leaves. In the proposed approach, the efficient convolutional layers are considered to extract the characteristics of each image sample. In the following, a layer is employed to classify the diseases. In this method, the transfer learning strategy is applied to extract the feature due to the small number of samples for the training phase. The advantages of using this model are summarized in the following:
	\begin{enumerate}
		\item Using layers of two super pre-trained models as a transfer learning strategy to better extract features;
		\item Presenting a compact structure made the training network easier and runnable in poor hardware.
		\item Due to the type of layer connectivity, the high and low-level features are concatenated in each layer, and the network's performance improves more and more.
		\item The efficiency of the Inception module at the end of the network, which extracts information at different levels, is another advantage of the proposed schemes.
	\end{enumerate}
	\subsection{Road map}
	The rest of this paper is organized as follows: Section \ref{sec:Proposed} briefly explains background material and the
	proposed method in detail. The experimental results, analysis, and performance evaluation are illustrated in Section \ref{sec:Experimental}. Finally, the conclusions and future scopes are introduced in Section \ref{sec:Conclusion}.
	
	\begin{table*}[t]
		\footnotesize
		\caption{The related parameters of the Mob-INC model.}
		\label{TABLE:param}
		\renewcommand{\arraystretch}{1.5}
		\scalebox{1} {
			\begin{tabular*}{\textwidth}{@{\extracolsep{\fill}}l@{}c@{}c@{}c@{}c}
				\cline{1-5}
				Module&Function&Output Shape&Expansion Factor&Stride\\
				\cline{1-5}
				Input&None&224$\times$224$\times$3&-&-\\
				Conv2d&Convolution&224$\times$224$\times$3&-&2\\
				Bottleneck&Depthwise,Convolution&112$\times$112$\times$32&1&1\\
				Bottleneck&Depthwise,Convolution&112$\times$112$\times$16&6&2\\
				Bottleneck&Depthwise,Convolution&56$\times$56$\times$24&6&2\\
				Bottleneck&Depthwise,Convolution&28$\times$28$\times$32&6&2\\
				Bottleneck&Depthwise,Convolution&14$\times$14$\times$64&6&1\\
				Bottleneck&Depthwise,Convolution&14$\times$14$\times$96&6&2\\
				Inception Module	&-&-&-&1\\
				Inception Module	&-&-&-&-\\
				FC1&	Fully connect&1$\times$1$\times$512&-&-\\
				Softmax&	Softmax Regression&1$\times$1$\times$4&-&-\\
				\cline{1-5}
		\end{tabular*}}
	\end{table*}
	\section{Proposed Method}
	\label{sec:Proposed}
	Deep learning has made great strides in recent years, especially in machine vision applications. Complex and challenging problems can effortlessly be solved with the help of these techniques. In contrast to the classic scheme, the deep network can extract features and classify instances simultaneously. However, the training time and the amount of data required for deep learning systems are much longer than traditional machine learning methods.
	
	Nowadays, this weakness has become a critical hot topic in deep learning. It allows deep neural networks to be trained with relatively little data. Actually, the effective trained models with insufficient data are exceptionally invaluable. There are a few labeled data to learn complex deep learning models in most real-world problems. In such hardships, the usage of established DCNNs, including pre-trained ResNet, Vgg-19, Mobilenet, InceptionV3, and DenseNet on a huge labeled natural image database, has proven to resolve cross-domain image classification problems with the help of transfer learning. 
	
	Generally, Convolution Neural Networks often try to find edges in the primary layers, shapes in the middle layers, and some dense high-level features in the final layers. In transitional learning, some of the mid and early layers of the model are retained, and only the last layers need to be retrained to make the model suitable for the new task. In other words, it is one of the advantages of transfer learning which can be employed without massive data. In summary, the concept of transfer learning is cheaper and more efficient. The transfer learning mechanism using any of the above-discussed models is briefly described in Figure \ref{fig:tl}.
	
	According to the stated points, this study presents a new architecture based on the previous structures by examining the popular models for disease recognition of maize leaves. In the following, the details of these structures are described, and then the proposed scheme is presented.
	\begin{figure*}[t]
		\center
		\begin{tabular}{cccc}
			\includegraphics[width=0.18\textwidth]{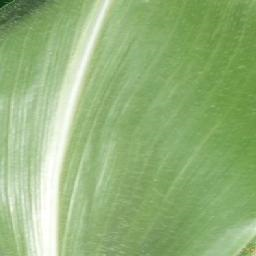} &
			\includegraphics[width=0.18\textwidth]{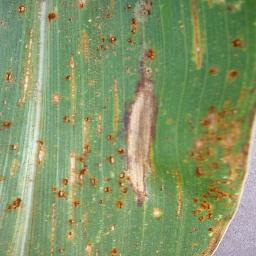} &
			\includegraphics[width=0.18\textwidth]{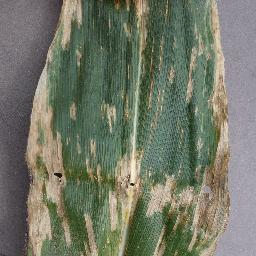} &
			\includegraphics[width=0.18\textwidth]{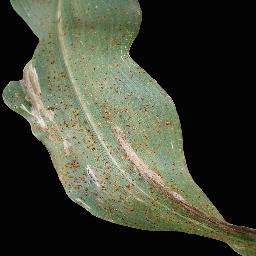} \\
			\includegraphics[width=0.18\textwidth]{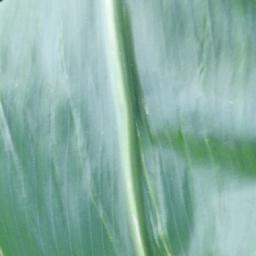} &
			\includegraphics[width=0.18\textwidth]{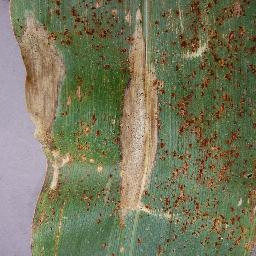} &
			\includegraphics[width=0.18\textwidth]{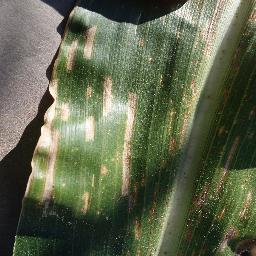} &
			\includegraphics[width=0.18\textwidth]{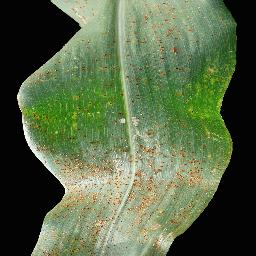} \\
			\includegraphics[width=0.18\textwidth]{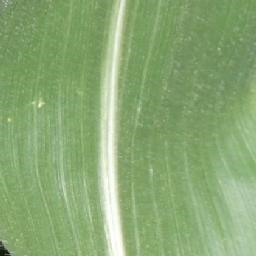} &
			\includegraphics[width=0.18\textwidth]{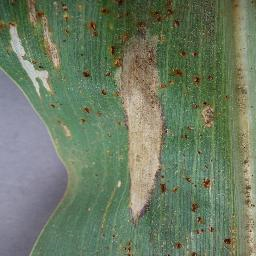} &
			\includegraphics[width=0.18\textwidth]{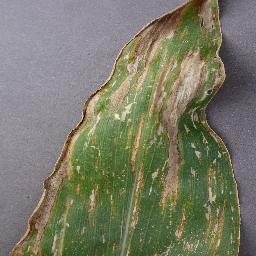} &
			\includegraphics[width=0.18\textwidth]{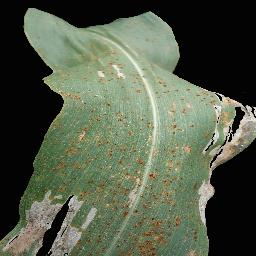} \\
			(a) & (b) & (c) & (d)
		\end{tabular}
		\caption{Some sample of the Plantvillage database\cite{ref31}, (a) Healthy, (b) Northern Leaf Blight, (c) Gray Leaf Spot, (d) Common Rust.}		
		\label{fig:dbdb}
	\end{figure*}
	\subsection{MobileNet}
	The mobile network has been developed to improve the real-time performance of deep neural networks under limited hardware conditions \cite{ref32}; It is practically one of the group of efficient torsional neural networks. This network can reduce the number of parameters without compromising accuracy. The basic idea behind MobileNet is that convolution layers can be replaced with separable depthwise and reduce the processing overhead. 
	
	In a typical convolution operation, each filter is applied to all input matrices, and then all the results are combined. In the MobileNet, instead of this mechanism, the convolution step is converted into two stages. First, a filter is applied to the matrix or feature mapping, and in the second step, a 1$\times$1 convolution layer is placed, which combines the matrices. After each convolution layer, a manual normalization algorithm and a ReLU activation function are employed, respectively. Totally, Mobilenet architecture is a set of blocks constructed with a depth and point convolution, batch normalization, and activation function. The complete mobile architecture includes a typical 3$\times$3 convolution layer as the first layer, followed by a set of blocks, and the last layer is a global max-pooling layer. The MobileNet also has a width factor parameter(alpha) as a depth multiplier that allows you to change the number of channels in each layer. With the help of the alpha coefficient, the network width and the number of parameters compared to the original version of the network can be customized.
	
	In this research, the latest version of the MobileNet as MobileNet V2 has been used. In the second version of MobileNet, there are three convolution layers in each block. Simultaneously, the main goal is to reduce the volume of computations and improve accuracy. The depth convolution layer filters the input, and the number of points of the channels is deduced using the pointwise strategy. Generally, these operations are named projection layers. In other words, it converts high-dimensional data (different channels) to lower-dimensional. Due to the lower parameter, this version is faster than the previous one. For more info about MobileNet refer to \cite{ref32}.

	\subsection{Inception}
	GoogleNet is an architecture designed by Google researchers which won ImageNet 2014 challenge. In addition to more depth (it has 22 layers, compared to VGG, which has 19 layers), the researchers also proposed a new approach called the Inception Module in this architecture. This module is a significant change from previous architectures. Several types of feature extractors layers receive input values and convert them into data for computation. This layering indirectly contributes to better network performance in a self-learning network that uses different options to solve tasks. This module can use inputs directly in its calculations or summarize them directly.
	
	The final architecture consists of several initial modules stacked on top of each other. The training process is different from the mentioned architectures because most top layers have their own output layer. This slight modification makes the learning process of this model faster because there is both general learning and parallel learning in layers\cite{ref34}. The advantages of GoogleNet compared to VGG can be summarized below:
	\begin{enumerate}
		\item GoogleNet is faster than VGG
		\item The number of layers of a pre-trained GoogleNet model is much lighter than a VGG. 
		\item A VGG model can be more than 500 MB in size, while GoogleNet is only 96 MB in size.
	\end{enumerate}
	
	Although large convolutional networks have acceptable performance, they are prone to overfitting due to many parameters. The computations will increase similarly by making the network uniform in size like VGG. On the other hand, by tending the opposite way as shallower networks, although theoretically usable, the data structures are absolutely non-optimal in practical situations. Hence, the researchers tried to approximate the thin design without the mentioned problems. Their idea was convolution filters with different sizes which could cover different parts of the information. For this aim, the different layers with varying sizes of the filter were created. In this way, the outputs of each filter are merged after extracting features and pass through the next layer. Accordingly, the performance can be increased 2 to 3 times compared to a network without these modules.
	
	\begin{table}[t]
		\footnotesize
		\caption{Th evaluation indicators of maize disease detection(\%).}
		\label{TABLE:acc}
		\renewcommand{\arraystretch}{1.5}
		\scalebox{1} {
			\begin{tabular*}{\columnwidth}{@{\extracolsep{\fill}}l@{}c@{}c@{}c@{}c}
				\cline{1-5}
				Type&Precision&Sensitivity&F1-Score&Specificity\\
				\cline{1-5}
				Gray Leaf Spot&0.89&	0.97&	0.93&	95.47\\
				Common Rust&0.99&	0.94&	0.96&	99.57\\
				Northern Leaf Blight&0.97&	0.92&	0.94&	98.00\\
				Healthy&1.00&	1.00&	1.00&	99.57\\
				\cline{1-5}
		\end{tabular*}}
	\end{table}
	
	
	\begin{figure*}[t]
		\center
		\begin{tabular}{@{}c@{}c}
			\includegraphics[width=0.43\textwidth]{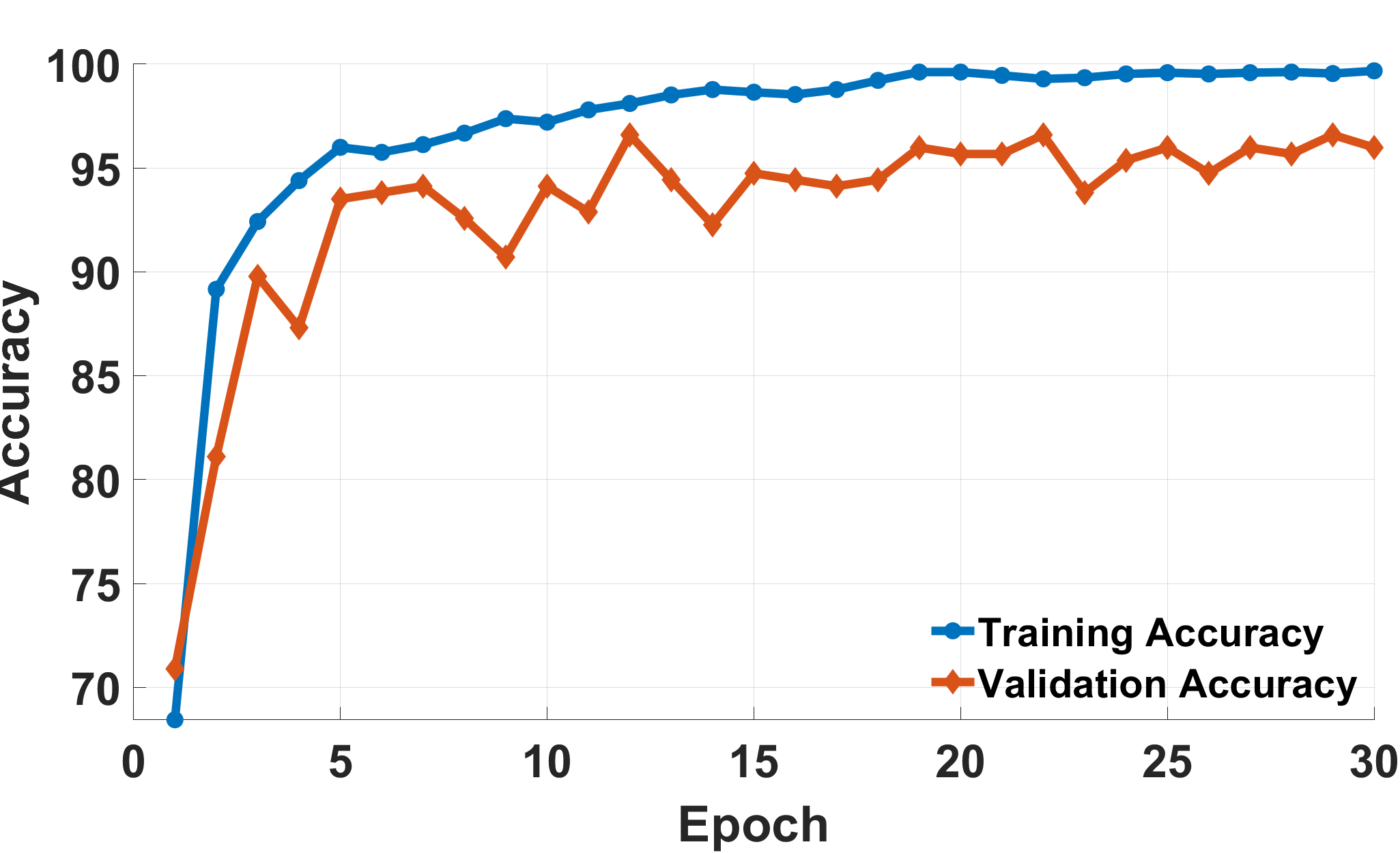} &
			\includegraphics[width=0.43\textwidth]{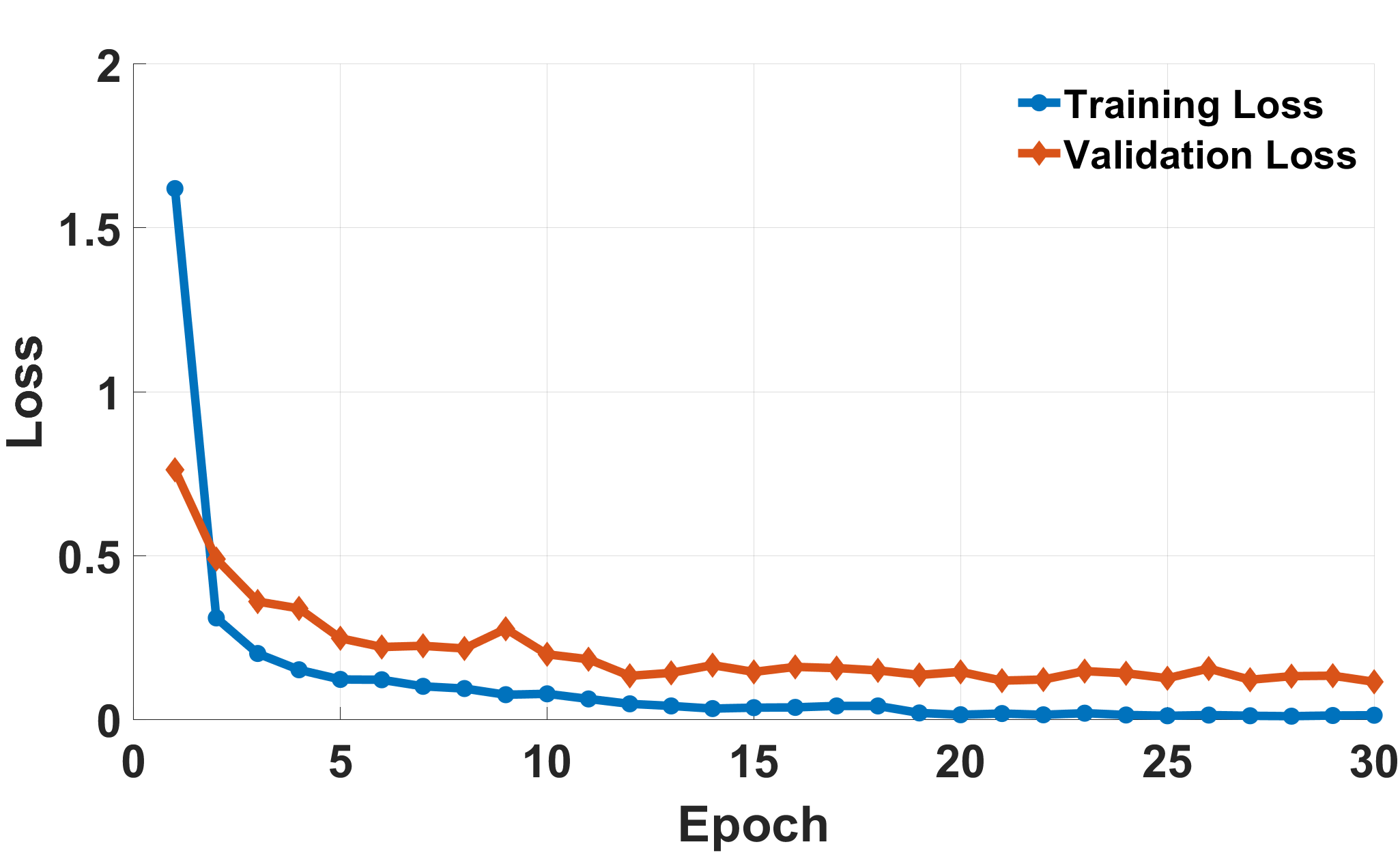} \\
			(a) &(b)\\
		\end{tabular}
		\caption{The performance of the proposed method in terms of accuracy and loss. (a) Shows the accuracy, and (b) Depicts the loss of the model for both train and validation data.}
		\label{fig:acc}
	\end{figure*}

	\begin{figure}[t!]
		\centering
		\includegraphics[width=1\columnwidth,trim= 0cm 0cm 0cm 0cm,clip]{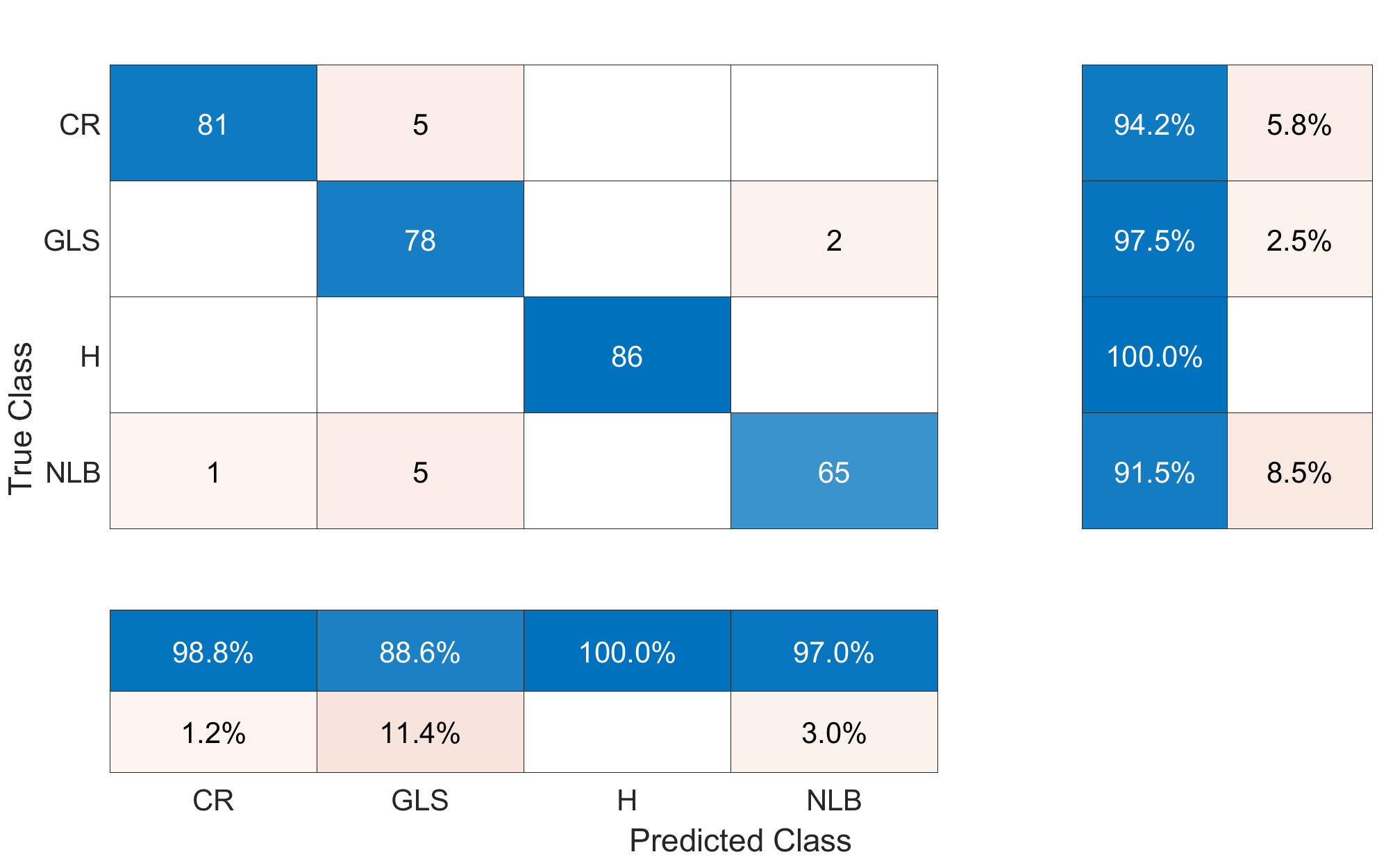}
		\caption{The confusion matrix of plant disease detection. (H) Healthy, (NLB) Northern Leaf Blight, (GLS) Gray Leaf Spot, (CR) Common Rust.}
		\label{fig:confusion}
	\end{figure}
	\subsection{Mob-INC}
	This subsection proposes a practical model named Mob-INC based on MobileNet and Inception. In detail, due to networks' efficiency and remarkable performance, a new scheme by uniting them is introduced to recognize the disease in the leaves of the maize plant. To combine these networks, first, the output of the 16th expand-block is given to the input of the inception module. Afterward, two inception modules are considered and connected to two fully connected layers.
	
	In this way, the first part, which belongs to MobileNet, is used to extract features, and the second part is an extended structure that focuses on multi-scale feature extraction. Generally, the presented network is performed by applying the following three steps:
	\begin{enumerate}
		\item Firstly, the training and testing data are scaled to a 224$\times$224 in the preprocessing phase. The distribution of data in the dataset for Gray Leaf Spot is unbalanced, so data augmentation is employed to solve this problem. In detail, the techniques used for data augmentation include horizontal or vertical flipping, rotating, shearing, crop, translation, etc. Moreover, the resampling mechanism is applied to tackle this challenge in the candidate dataset.
		\item In the second step, the pre-trained MobileNetV2 on ImageNet is customized. For this aim, the input layers and fully connected layers are modified.
		\item Next, the network is configured to categorize the new class. The weight of MobileNet's layers is frozen to the last six layers, and then training is done on the remaining layers.
		\item In the last step, the original model is expanded by adding two layers of the Inception module and a fully connected layer with 512 neurons. All layers of the new model are trained on new image sets. Notice that the Softmax and Categorical Cross-Entropy are utilized as active and loss functions, respectively.
	\end{enumerate}
	
	Figure \ref{fig:diagram} shows the network structure and the corresponding parameters are presented in Table \ref{TABLE:param}.

	\begin{table*}[t]
		\footnotesize
		\caption{The training results of different approaches on the public rice dataset}
		\label{TABLE:accn}
		\renewcommand{\arraystretch}{1.5}
		\scalebox{1} {
			\begin{tabular*}{\textwidth}{@{\extracolsep{\fill}}l@{}c@{}c@{}c@{}c@{}c@{}c@{}c@{}c}
				\cline{1-9}
				&\multicolumn{4}{c}{10 Epochs}&\multicolumn{4}{c}{30 Epochs}\\
				\cline{2-5}\cline{5-9}
				&\multicolumn{2}{c}{Accuracy}&\multicolumn{2}{c}{Loss}&\multicolumn{2}{c}{Accuracy}&\multicolumn{2}{c}{Loss}\\
				\cline{2-3}\cline{4-5}\cline{6-7}\cline{8-9}
				Model&Training &Validation &Training &Validation &	Training &Validation &Training &Validation \\
				\cline{1-9}
				VGG-19&89.27&89.67&1.7341&2.056&92.65&90.34&0.6506&0.7124\\
				Resnet-50&85.89&83.54&0.8993&1.058&89.78&86.56&0.7534&0.8066\\
				Inceptionv3&85.76&90.76&0.6169&0.8945&93.67&92.98&04803&0.6119\\
				Densenet201&85.78&84.44&0.3879&0.4892&99.31&93.17&0.9856&0.2555\\
				Mobilenet&88.65&85.23&0.5985&0.7845&97.34&93.18&0.1246&0.2681\\
				Mob-INC&95.67&93.56&0.1425&0.2346&99.72&96.10&0.01&0.1466\\
				\cline{1-9}
		\end{tabular*}}
	\end{table*}
	\begin{figure*}[t]
		\center
		\begin{tabular}{@{}c@{}c@{}c@{}c@{}}

			\includegraphics[width=0.24\textwidth]{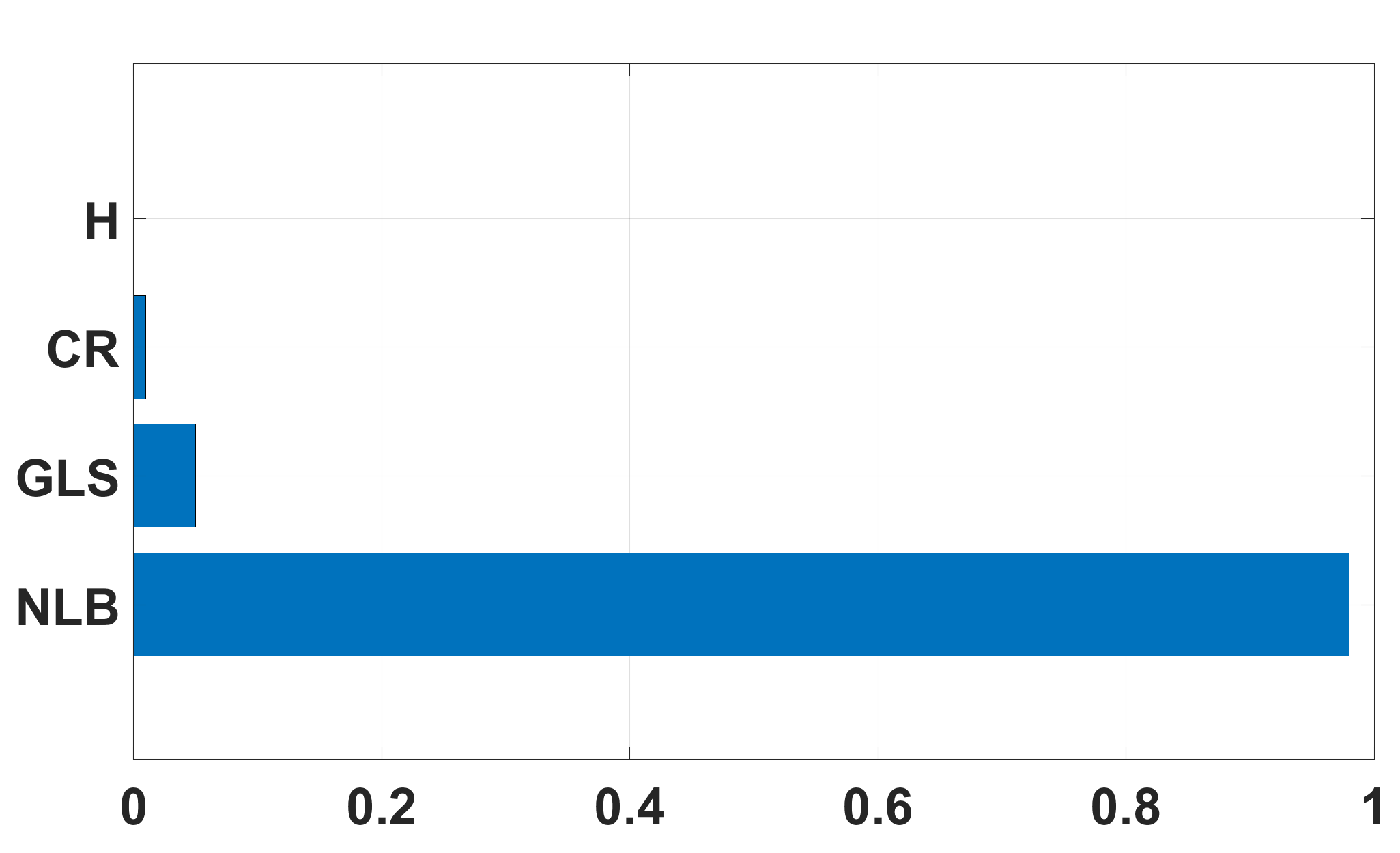} &
			\includegraphics[width=0.24\textwidth]{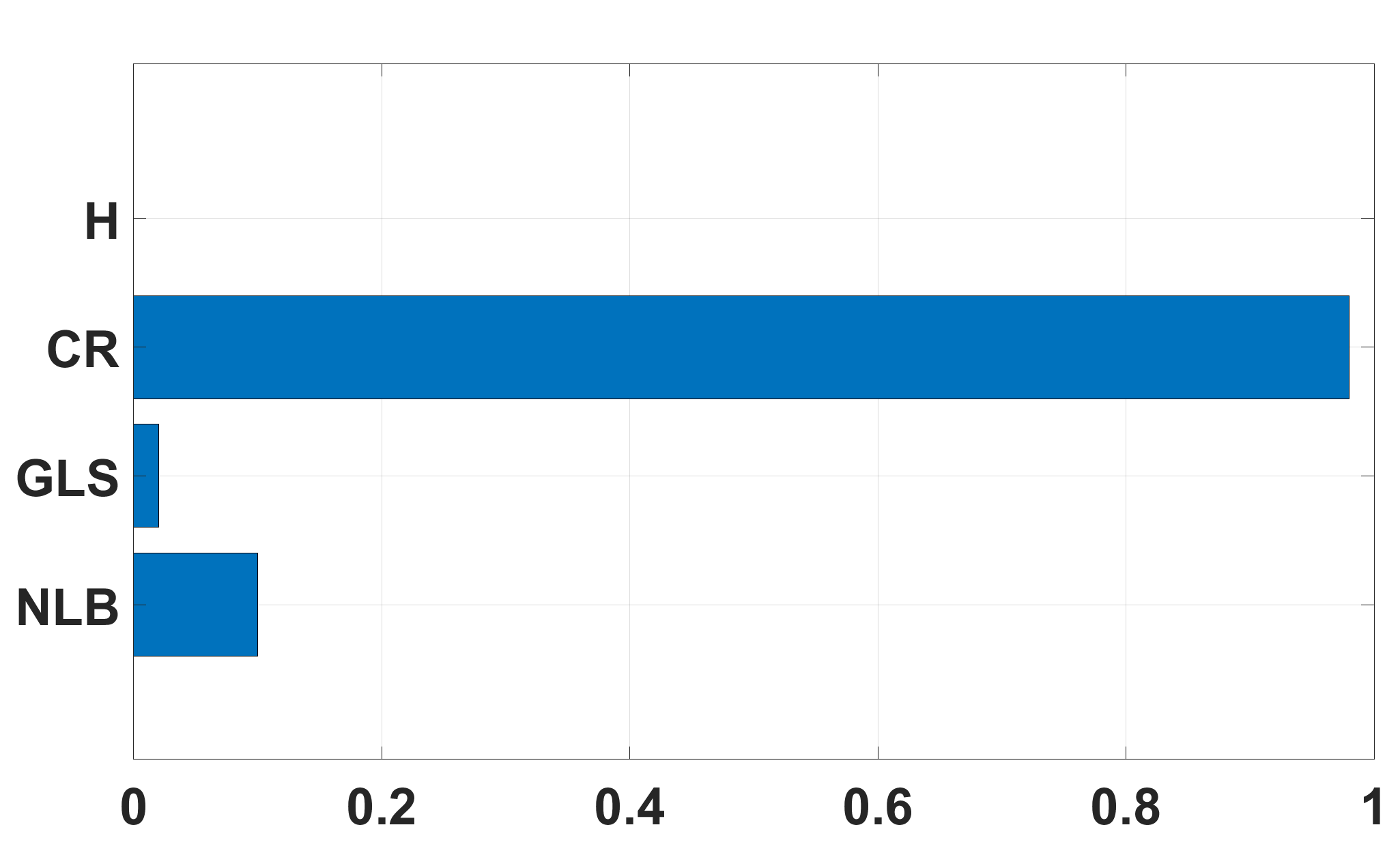} &
			\includegraphics[width=0.24\textwidth]{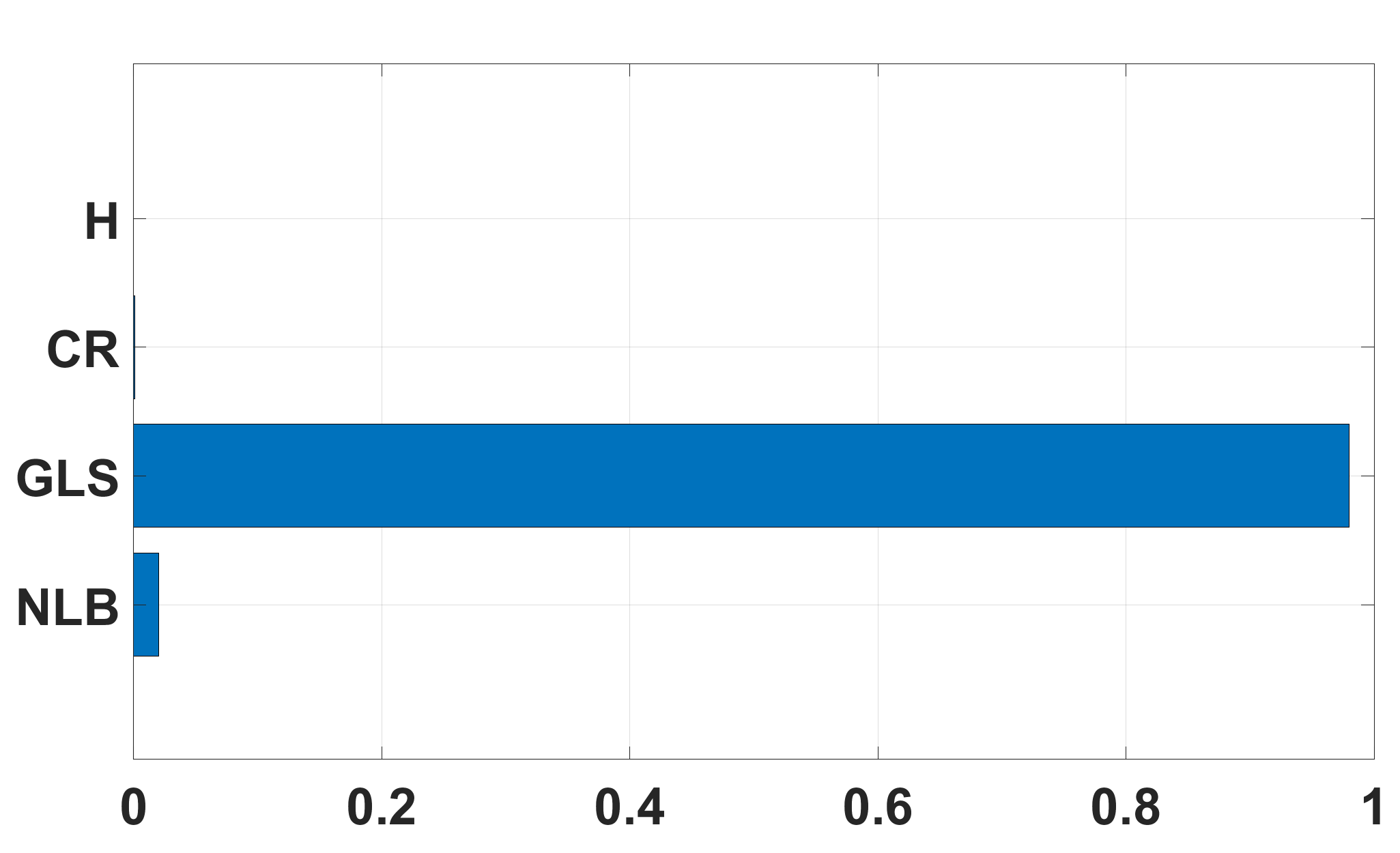} &
			\includegraphics[width=0.24\textwidth]{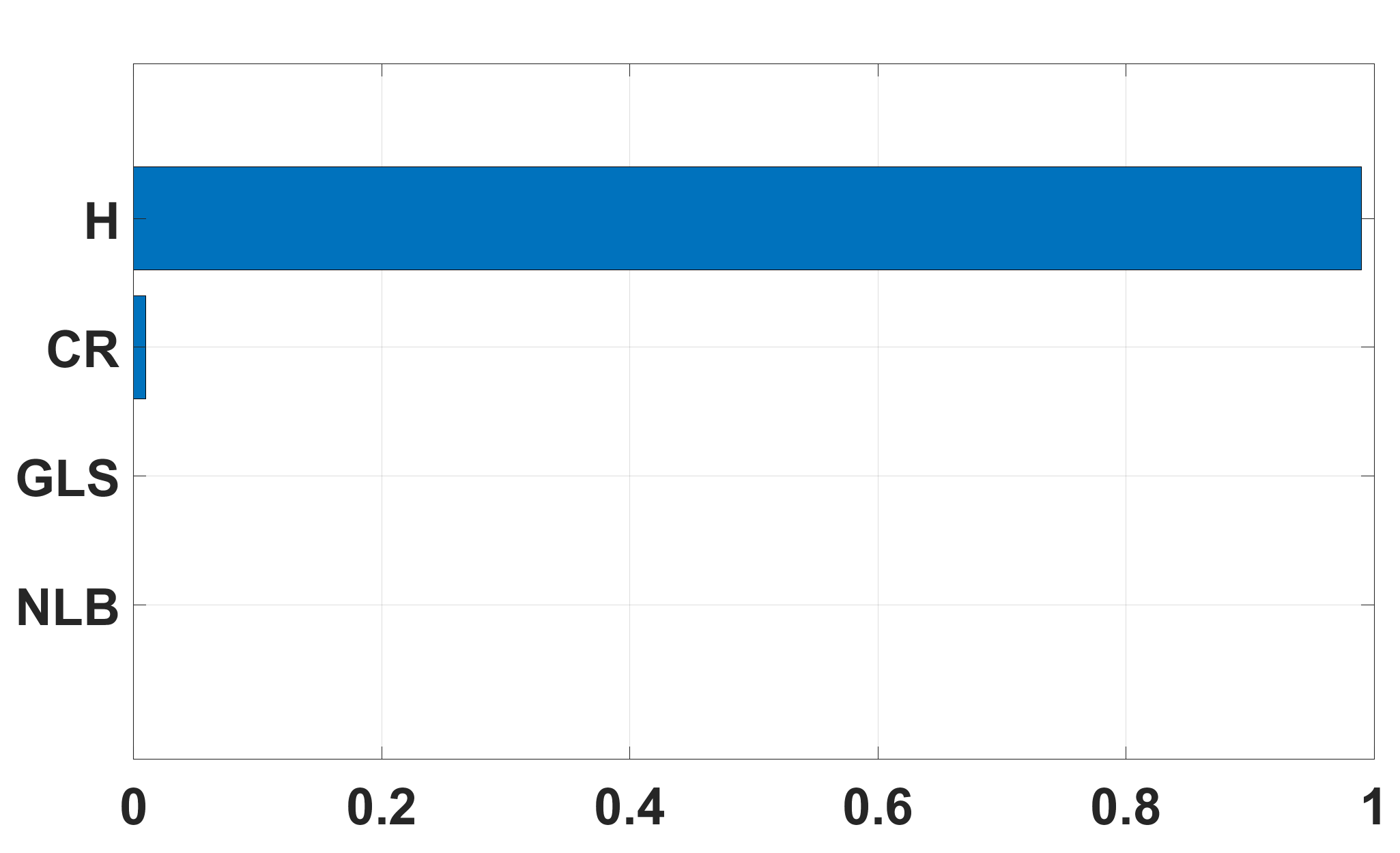} \\
			
			\includegraphics[width=0.23\textwidth]{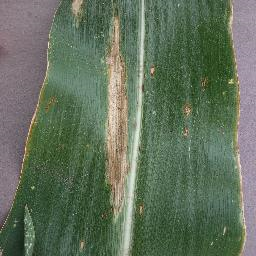} &
			\includegraphics[width=0.23\textwidth]{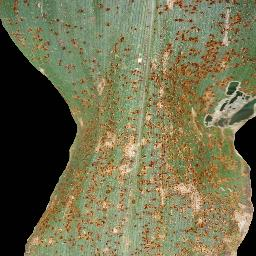} &
			\includegraphics[width=0.23\textwidth]{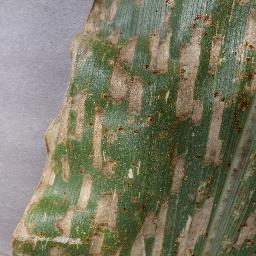} &
			\includegraphics[width=0.23\textwidth]{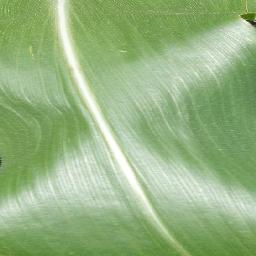} \\
			(a)&(b)&(c)&(d)\\
		\end{tabular}
		\caption{The classification output of plant disease for each class. (a) Northern Leaf Blight, (b) Common Rust, (c) Gray Leaf Spot, and (d) Healthy.}
		\label{fig:prob}
	\end{figure*}
	
	\section{Experimental Results}
	\label{sec:Experimental}
	In the present work, data comes from the Plant village dataset \cite{ref31}. This is a general international dataset for plant disease detection. The dataset has leaf images of 14 plants with 38 diseases and contains 54,306 images of diseased and healthy plant leaves. The maize leaf subset of this dataset was used in this paper. The mentioned dataset has 3,852 color images categorized into four groups, including 513 gray leaf spot images, 1,192 common rust images, 1,162 healthy images, and 985 northern leaf blight images. The whole image size is 256$\times$256, all of which are in color space mode. All experiments were implemented on a 64-bit system equipped with GPU 1070 Ti. The deep learning-based models have also been implemented on Tensorflow 2.3.0 and Python 3.7.
	
	First of all, the definition of these diseases are briefly summarized below:
	
	The Common Rust (CR) disease reduces the yield of the plan by decreasing the photosynthetic level of the leaves. If the disease starts before coagulation, it causes significant damage to grain yield and is named Northern Leaf Blight (NLB). The cause of this disease is Helminthosporium turcicum. As the last disease, the Gray Leaf Spot (GLS) is a foliar fungal disease that affects maize. It is one of the most effective yield-limiting diseases of maize worldwide. Some samples of this dataset for each category are demonstrated in Figure \ref{fig:dbdb}. The number of classes and various textures, including edge, smooth, and rough, in this dataset, challenges the proposed leaves disease detection problem models.

	In the following, a series of experiments are reported to prove the superiority and efficiency of Mob-INC compared to other state-of-the-art models.
	
	First, the accuracy and loss of training and validation phases are illustrated in Figure \ref{fig:acc}. By looking at details, it can be found that the network's performance reaches an admissible value after 15 epochs. Also, the accuracy of the validation data is slightly lower than the train. Based on the performed experiments, the epoch number equal to 30 is suitable for training the model. Something else which should be mentioned here is that in the training phase, 70\% of the images are considered for training, 10\% are employed for validation, and the rest are utilized for the test phase; The efficiency of this partitioning is proved in \cite{ref19}.
	
	Next, in Table \ref{TABLE:acc}, five known metrics, including Precision, F1-Score, Sensitivity, and Specificity, are reported to indicate the accuracy of the scheme. In addition, the confusion matrix of model is demonstrated in Figure \ref{fig:confusion}. For this aim, a set of images is selected to predict each class and the results of this prediction. As seen, the model can classify healthy leaves without any problems. On the other hand, for three main classes, some mistakes have been observed in Figure \ref{fig:confusion}. Obviously, the performance evaluation of the proposed model based on the mentioned terms significantly proves the efficiency of Mob-INC.
	
	
	In the following, a probability of belonging to each class for three primary diseases is illustrated in Figure \ref{fig:prob}. It is evident in these cases that the model can genuinely classify images with high probability without any overlap with the rest classes. In other words, the classification layer of the model can exactly recognize the label of images.
	
	Lastly, in Table \ref{TABLE:accn}, the accuracy and loss of taring and validation phases are compared to state-of-the-art networks, which were presented in recent years to prove the efficiency of Mob-INC. As can be seen, the proposed model has absolute superiority and outperforms the mentioned models. This improvement is due to combining architectures and employing a pre-trained network. Hence, the scheme can profit from the advantages of both models. By looking in detail, it can be found that most plans after 30 epochs can not touch the reached accuracy by Mob-INC when the epoch is set to 10. After 30 epochs, the accuracy and loss of validation data of the proposed models reach 96.10\% and 0.14, respectively.
	
	In brief, the results reveal that the proposed model demonstrates high accuracy for maize disease detection. In other words, it has a significant capability to recognize the different diseases and extend for real-time applications.
	
	\section{Conclusion and Future Works}
	\label{sec:Conclusion}
	Rapid diagnosis of plant diseases has always been one of the main challenges of the agricultural industry. Plant diseases impact food safety and can seriously reduce the quality and quantity of agricultural products. Besides, there are many challenges to accurate and timely diagnosis of the disease. According to previous research, it has been studied that deep learning has an outstanding performance in the detection of leaf disease. Therefore, a novel architecture named MobINC using transfer learning was presented in this work. In this method, the pre-trained MobileNetV2 model has been used. First, the model has been fine-tuned to be usable for the intended application. For this purpose, whole layers were employed without modification. In the following, an Inception module was appended to the network and then presented to a new fully-connected layer. These modifications lead to improving the accuracy and efficiency of the network. Briefly, experimental results indicate the superiority of the proposed method compared to the rest of the state-of-the-art models. 
	
	The work will be extended to examine the instance labels with different backgrounds in the future. Moreover, it will be developed for poor devices so that farmers can treat them with a timely diagnosis of the disease.

\begin{thebibliography}{10}
\expandafter\ifx\csname url\endcsname\relax
  \def\url#1{\texttt{#1}}\fi
\expandafter\ifx\csname urlprefix\endcsname\relax\def\urlprefix{URL }\fi
\expandafter\ifx\csname href\endcsname\relax
  \def\href#1#2{#2} \def\path#1{#1}\fi

\bibitem{ref1}
X.~Li, L.~Zhang, Z.~Fu, X.~Liu, H.~Wen, et~al., The corn disease remote
  diagnostic system in china., Journal of Food, Agriculture \& Environment
  10~(1 part 2) (2012) 617--620.

\bibitem{ref2}
S.~Bagde, S.~Patil, S.~Patil, P.~Patil, Artificial neural network based plant
  leaf disease detection, IJCSMC 4~(4) (2015) 900--905.

\bibitem{ref3}
V.~Pooja, R.~Das, V.~Kanchana, Identification of plant leaf diseases using
  image processing techniques, in: 2017 IEEE Technological Innovations in ICT
  for Agriculture and Rural Development (TIAR), 2017, pp. 130--133.
\newblock \href {https://doi.org/10.1109/TIAR.2017.8273700}
  {\path{doi:10.1109/TIAR.2017.8273700}}.

\bibitem{ref4}
X.~Bai, Z.~Cao, L.~Zhao, J.~Zhang, C.~Lv, C.~Li, J.~Xie,
  \href{https://www.sciencedirect.com/science/article/pii/S0168192318301473}{Rice
  heading stage automatic observation by multi-classifier cascade based rice
  spike detection method}, Agricultural and Forest Meteorology 259 (2018)
  260--270.
\newblock \href
  {https://doi.org/https://doi.org/10.1016/j.agrformet.2018.05.001}
  {\path{doi:https://doi.org/10.1016/j.agrformet.2018.05.001}}.
\newline\urlprefix\url{https://www.sciencedirect.com/science/article/pii/S0168192318301473}

\bibitem{ref5}
K.~P. Ferentinos,
  \href{https://www.sciencedirect.com/science/article/pii/S0168169917311742}{Deep
  learning models for plant disease detection and diagnosis}, Computers and
  Electronics in Agriculture 145 (2018) 311--318.
\newblock \href {https://doi.org/https://doi.org/10.1016/j.compag.2018.01.009}
  {\path{doi:https://doi.org/10.1016/j.compag.2018.01.009}}.
\newline\urlprefix\url{https://www.sciencedirect.com/science/article/pii/S0168169917311742}

\bibitem{ref6}
J.~Chen, J.~Chen, D.~Zhang, Y.~Sun, Y.~Nanehkaran,
  \href{https://www.sciencedirect.com/science/article/pii/S0168169919322422}{Using
  deep transfer learning for image-based plant disease identification},
  Computers and Electronics in Agriculture 173 (2020) 105393.
\newblock \href {https://doi.org/https://doi.org/10.1016/j.compag.2020.105393}
  {\path{doi:https://doi.org/10.1016/j.compag.2020.105393}}.
\newline\urlprefix\url{https://www.sciencedirect.com/science/article/pii/S0168169919322422}

\bibitem{ref7}
\href{https://www.sciencedirect.com/science/article/pii/S2214317317301774}{A
  review of neural networks in plant disease detection using hyperspectral
  data}, Information Processing in Agriculture 5~(3) (2018) 354--371.
\newblock \href {https://doi.org/https://doi.org/10.1016/j.inpa.2018.05.002}
  {\path{doi:https://doi.org/10.1016/j.inpa.2018.05.002}}.
\newline\urlprefix\url{https://www.sciencedirect.com/science/article/pii/S2214317317301774}

\bibitem{ref8}
J.~Ma, K.~Du, F.~Zheng, L.~Zhang, Z.~Gong, Z.~Sun,
  \href{https://www.sciencedirect.com/science/article/pii/S0168169918309360}{A
  recognition method for cucumber diseases using leaf symptom images based on
  deep convolutional neural network}, Computers and Electronics in Agriculture
  154 (2018) 18--24.
\newblock \href {https://doi.org/https://doi.org/10.1016/j.compag.2018.08.048}
  {\path{doi:https://doi.org/10.1016/j.compag.2018.08.048}}.
\newline\urlprefix\url{https://www.sciencedirect.com/science/article/pii/S0168169918309360}

\bibitem{ref9}
J.~Garc{\'\i}a, C.~Pope, F.~Altimiras, A distributed-means segmentation
  algorithm applied to lobesia botrana recognition, Complexity 2017 (2017).

\bibitem{ref10}
N.~Guettari, A.~S. Capelle-Laizé, P.~Carré, Blind image steganalysis based on
  evidential k-nearest neighbors, in: 2016 IEEE International Conference on
  Image Processing (ICIP), 2016, pp. 2742--2746.
\newblock \href {https://doi.org/10.1109/ICIP.2016.7532858}
  {\path{doi:10.1109/ICIP.2016.7532858}}.

\bibitem{ref11}
S.~Deepa, R.~Umarani, Steganalysis on images using svm with selected hybrid
  features of gini index feature selection algorithm., International Journal of
  Advanced Research in Computer Science 8~(5) (2017).

\bibitem{ref12}
M.~Ramezani, S.~Ghaemmaghami, Towards genetic feature selection in image
  steganalysis, in: 2010 7th IEEE Consumer Communications and Networking
  Conference, 2010, pp. 1--4.
\newblock \href {https://doi.org/10.1109/CCNC.2010.5421805}
  {\path{doi:10.1109/CCNC.2010.5421805}}.

\bibitem{ref13}
J.~Kodovsky, J.~Fridrich, V.~Holub, Ensemble classifiers for steganalysis of
  digital media, IEEE Transactions on Information Forensics and Security 7~(2)
  (2012) 432--444.
\newblock \href {https://doi.org/10.1109/TIFS.2011.2175919}
  {\path{doi:10.1109/TIFS.2011.2175919}}.

\bibitem{ref14}
Y.~Guo, T.~Hastie, R.~Tibshirani, Regularized linear discriminant analysis and
  its application in microarrays, Biostatistics 8~(1) (2007) 86--100.

\bibitem{ref15}
S.~Zhang, Z.~Wang,
  \href{https://www.sciencedirect.com/science/article/pii/S092523121630296X}{Cucumber
  disease recognition based on global-local singular value decomposition},
  Neurocomputing 205 (2016) 341--348.
\newblock \href {https://doi.org/https://doi.org/10.1016/j.neucom.2016.04.034}
  {\path{doi:https://doi.org/10.1016/j.neucom.2016.04.034}}.
\newline\urlprefix\url{https://www.sciencedirect.com/science/article/pii/S092523121630296X}

\bibitem{ref16}
\href{https://www.sciencedirect.com/science/article/pii/S0168169917300820}{Leaf
  image based cucumber disease recognition using sparse representation
  classification}, Computers and Electronics in Agriculture 134 (2017)
  135--141.
\newblock \href {https://doi.org/https://doi.org/10.1016/j.compag.2017.01.014}
  {\path{doi:https://doi.org/10.1016/j.compag.2017.01.014}}.
\newline\urlprefix\url{https://www.sciencedirect.com/science/article/pii/S0168169917300820}

\bibitem{ref22}
A.~Kamilaris, F.~X. Prenafeta-Boldú,
  \href{https://www.sciencedirect.com/science/article/pii/S0168169917308803}{Deep
  learning in agriculture: A survey}, Computers and Electronics in Agriculture
  147 (2018) 70--90.
\newblock \href {https://doi.org/https://doi.org/10.1016/j.compag.2018.02.016}
  {\path{doi:https://doi.org/10.1016/j.compag.2018.02.016}}.
\newline\urlprefix\url{https://www.sciencedirect.com/science/article/pii/S0168169917308803}

\bibitem{ref17}
Y.~LeCun, B.~Boser, J.~S. Denker, D.~Henderson, R.~E. Howard, W.~Hubbard, L.~D.
  Jackel, Backpropagation applied to handwritten zip code recognition, Neural
  Computation 1~(4) (1989) 541--551.
\newblock \href {https://doi.org/10.1162/neco.1989.1.4.541}
  {\path{doi:10.1162/neco.1989.1.4.541}}.

\bibitem{ref18}
\href{https://www.sciencedirect.com/science/article/pii/S0925231217311384}{Identification
  of rice diseases using deep convolutional neural networks}, Neurocomputing
  267 (2017) 378--384.
\newblock \href {https://doi.org/https://doi.org/10.1016/j.neucom.2017.06.023}
  {\path{doi:https://doi.org/10.1016/j.neucom.2017.06.023}}.
\newline\urlprefix\url{https://www.sciencedirect.com/science/article/pii/S0925231217311384}

\bibitem{ref19}
S.~P. Mohanty, D.~Hughes, M.~Salathe, Inference of plant diseases from leaf
  images through deep learning, Front. Plant Sci 7 (2016) 1419.

\bibitem{ref20}
C.~DeChant, T.~Wiesner-Hanks, S.~Chen, E.~L. Stewart, J.~Yosinski, M.~A. Gore,
  R.~J. Nelson, H.~Lipson, Automated identification of northern leaf
  blight-infected maize plants from field imagery using deep learning,
  Phytopathology 107~(11) (2017) 1426--1432.

\bibitem{ref21}
S.~P. Mohanty, D.~P. Hughes, M.~Salath{\'e}, Using deep learning for
  image-based plant disease detection, Frontiers in plant science 7 (2016)
  1419.

\bibitem{ref23}
X.~Zhang, Y.~Qiao, F.~Meng, C.~Fan, M.~Zhang, Identification of maize leaf
  diseases using improved deep convolutional neural networks, IEEE Access 6
  (2018) 30370--30377.
\newblock \href {https://doi.org/10.1109/ACCESS.2018.2844405}
  {\path{doi:10.1109/ACCESS.2018.2844405}}.

\bibitem{ref24}
\href{https://www.sciencedirect.com/science/article/pii/S1877050918310159}{Tomato
  crop disease classification using pre-trained deep learning algorithm},
  Procedia Computer Science 133 (2018) 1040--1047, international Conference on
  Robotics and Smart Manufacturing (RoSMa2018).
\newblock \href {https://doi.org/https://doi.org/10.1016/j.procs.2018.07.070}
  {\path{doi:https://doi.org/10.1016/j.procs.2018.07.070}}.
\newline\urlprefix\url{https://www.sciencedirect.com/science/article/pii/S1877050918310159}

\bibitem{ref25}
E.~C. Too, L.~Yujian, S.~Njuki, L.~Yingchun,
  \href{https://www.sciencedirect.com/science/article/pii/S0168169917313303}{A
  comparative study of fine-tuning deep learning models for plant disease
  identification}, Computers and Electronics in Agriculture 161 (2019)
  272--279, bigData and DSS in Agriculture.
\newblock \href {https://doi.org/https://doi.org/10.1016/j.compag.2018.03.032}
  {\path{doi:https://doi.org/10.1016/j.compag.2018.03.032}}.
\newline\urlprefix\url{https://www.sciencedirect.com/science/article/pii/S0168169917313303}

\bibitem{ref26}
G.~Hu, H.~Wu, Y.~Zhang, M.~Wan,
  \href{https://www.sciencedirect.com/science/article/pii/S0168169919300407}{A
  low shot learning method for tea leaf’s disease identification}, Computers
  and Electronics in Agriculture 163 (2019) 104852.
\newblock \href {https://doi.org/https://doi.org/10.1016/j.compag.2019.104852}
  {\path{doi:https://doi.org/10.1016/j.compag.2019.104852}}.
\newline\urlprefix\url{https://www.sciencedirect.com/science/article/pii/S0168169919300407}

\bibitem{ref27}
J.~G. {Arnal Barbedo},
  \href{https://www.sciencedirect.com/science/article/pii/S1537511018307797}{Plant
  disease identification from individual lesions and spots using deep
  learning}, Biosystems Engineering 180 (2019) 96--107.
\newblock \href
  {https://doi.org/https://doi.org/10.1016/j.biosystemseng.2019.02.002}
  {\path{doi:https://doi.org/10.1016/j.biosystemseng.2019.02.002}}.
\newline\urlprefix\url{https://www.sciencedirect.com/science/article/pii/S1537511018307797}

\bibitem{ref28}
J.~Arora, U.~Agrawal, et~al., Classification of maize leaf diseases from
  healthy leaves using deep forest, Journal of Artificial Intelligence and
  Systems 2~(1) (2020) 14--26.

\bibitem{ref29}
V.~V. Adit, C.~Rubesh, S.~S. Bharathi, G.~Santhiya, R.~Anuradha, A comparison
  of deep learning algorithms for plant disease classification, in: Advances in
  Cybernetics, Cognition, and Machine Learning for Communication Technologies,
  Springer, 2020, pp. 153--161.

\bibitem{ref30}
A.~Khamparia, G.~Saini, D.~Gupta, A.~Khanna, S.~Tiwari, V.~H.~C.
  de~Albuquerque, Seasonal crops disease prediction and classification using
  deep convolutional encoder network, Circuits, Systems, and Signal Processing
  39~(2) (2020) 818--836.

\bibitem{ref31}
D.~Hughes, M.~Salath{\'e}, et~al., An open access repository of images on plant
  health to enable the development of mobile disease diagnostics, arXiv
  preprint arXiv:1511.08060 (2015).

\bibitem{ref32}
M.~Sandler, A.~Howard, M.~Zhu, A.~Zhmoginov, L.-C. Chen, Mobilenetv2: Inverted
  residuals and linear bottlenecks, in: Proceedings of the IEEE conference on
  computer vision and pattern recognition, 2018, pp. 4510--4520.

\bibitem{ref34}
C.~Szegedy, S.~Ioffe, V.~Vanhoucke, A.~Alemi, Inception-v4, inception-resnet
  and the impact of residual connections on learning, arXiv preprint
  arXiv:1602.07261 (2016).

\end{thebibliography}

\end{document}